\documentclass[10pt,a4paper]{article}

\usepackage{cogsci}

\cogscifinalcopy % Uncomment this line for the final submission 

\usepackage{amssymb}
\usepackage{pslatex}
\usepackage{apacite}
\usepackage{multirow,makecell}
\usepackage{booktabs}
\usepackage{float} % Roger Levy added this and changed figure/table
\usepackage{amsmath}
\usepackage{linguex}
\usepackage{graphicx}
\usepackage{booktabs}
\usepackage[dvipsnames]{xcolor}

\title{
    Experimental Pragmatics with Machines: Testing LLM Predictions for the Inferences of Plain and Embedded Disjunctions
}
 
\author{
  {\large \bf Polina Tsvilodub (polina.tsvilodub@uni-tuebingen.de)} \\
  Department of Linguistics, University of Tübingen 
  \AND 
  {\large \bf Paul Marty (paul.pierre.marty@gmail.com)} \\
  Institute of Linguistics and Language Technology, University of Malta
  \AND 
  {\large \bf Sonia Ramotowska (ramotowska.or.s@gmail.com) } \\
  Institute for Logic, Language and Computation, University of Amsterdam \AND  
  {\large \bf Jacopo Romoli (jacopo.romoli@hhu.de)} \\
  Department of Linguistics, University of Düsseldorf
  \AND {\large \bf Michael Franke (michael.franke@uni-tuebingen.de)} \\
  Department of Linguistics, University of Tübingen 
  }

\begin{document}

\maketitle

\begin{abstract}
Human communication is based on a variety of inferences that we draw from sentences, often going beyond what is literally said. While there is wide agreement on the basic distinction between entailment, implicature, and presupposition, the status of many inferences remains controversial. In this paper, we focus on three inferences of plain and embedded disjunctions, and compare them with regular scalar implicatures. We investigate this comparison from the novel perspective of the predictions of state-of-the-art large language models, using the same experimental paradigms as recent studies investigating the same inferences with humans. The results of our best performing models mostly align with those of humans, both in the large differences we find between those inferences and implicatures, as well as in fine-grained distinctions among different aspects of those inferences. 

\end{abstract}

\section{Introduction}
Human communication is based on a variety of inferences we draw from sentences, often going beyond what is literally said \cite[and much subsequent work]{Grice1975:Logic-and-Conve}. One of the main results of theoretical approaches to meaning is the discovery of a rich typology of inferences, exhibiting different conversational status and divergent behavior in complex sentences. Over the last two decades, experimental work in pragmatics has tested the often subtle predictions of different theories of these inferences, by comparing them across a variety of tasks and measures with human participants \cite[and references therein]{noveck2018experimental}. However, while there is a general consensus regarding the basic differences between entailment, implicature, and presupposition, their boundaries are continuously being re-drawn, and the status of many inferences remains open. In recent years, advancements in machine learning have added the novel perspective of the investigation of these inferences in non-human language agents such as large language models (LLMs) in systematic comparison to humans \cite{GauthierHu2020:SyntaxGym:-An-O, warstadt-etal-2020-blimp-benchmark}. The integration of these different perspectives has increased our understanding of the typology mentioned above.

In this paper, we focus on three of such inferences, triggered by \emph{disjunction} (``or'') in different configurations, which we describe in turn. Firstly, plain disjunctions like \ref{Ig} are associated with so-called \textsc{ignorance} inferences (II), inferences that the speaker is uncertain as to which disjunct is true \Next[a] and  considers each of them possible \Next[b] \cite[among many others]{sauerland2004scalar}.

\ex.\label{Ig} This box contains a blue ball or a yellow ball.
\a.\label{Ig1} $\leadsto$ \textit{The speaker is not certain that the box contains a blue ball and s/he is not certain that it contains a yellow ball.} \hfill \textsc{uncertainty}
\b.\label{Ig2} $\leadsto$ \textit{The speaker deems it possible that the box contains a blue ball and that it contains a yellow ball.} \hfill \textsc{possibility}

Next, when embedded under a universal quantifier as in \ref{dist}, disjunctions can give rise to so-called \textsc{distributive} inferences (DI), meaning that the property associated with each disjunct applies to some but not all individuals in the domain of the quantifier \cite[among others]{sauerland2004scalar, Fox2007}. 

\ex.\label{dist} Every box contains a blue ball or a yellow ball. 
\a.\label{dist1}$\leadsto$ \textit{Not every box contains a blue ball and not every box contains a yellow ball.} \hfill \textsc{negated universal}
\b.\label{dist2}$\leadsto$ \textit{Some box contains a blue ball and some box contains a yellow ball.}\hfill\textsc{distributive}

Finally, in the scope of a possibility modal as in \ref{Ig3}, the resulting inference, called \textsc{free choice} (FC), conveys that each of the disjuncts is an open possibility \cite[among others]{Kamp1978, Fox2007}.  

\ex.\label{Ig3} This box might contain a blue ball or a yellow ball. \\ $\leadsto$ \textit{The box might contain a blue ball and it might contain a yellow ball.} \hfill\textsc{free~choice}

These inferences have all been argued to be derived as, or to result from scalar implicatures (SI) by some theories \cite[among others]{sauerland2004scalar, Fox2007, bar2020free} and thus, to rely on the same mechanisms as those deriving regular SIs like the one in \ref{Imp}.

\ex.\label{Imp} It is possible that this box contains a blue ball.\\
$\leadsto$ \textit{It is not certain that the box contains a blue ball.} \\ \textcolor{white}{x}  \hfill \textsc{scalar implicature}

Experimental data with human subjects has recently challenged the standard implicature approach to these inferences. First, \citeA{Marty:2021b} show that all these inferences are more readily derived by speakers than regular SIs, suggesting that they should be treated differently. Second, according to  traditional implicature accounts like \citeA{sauerland2004scalar}'s, \textsc{ignorance} and \textsc{distributive} inferences are derived in two consecutive steps. For IIs, the first step involves deriving the \textsc{uncertainty} (UNC) part in \ref{Ig1}, from which the \textsc{possibility} (POS) part in \ref{Ig2} follows; for DIs, the first step involves deriving the \textsc{negated universal} (NU) part in \ref{dist1}, from which the \textsc{distributive} (DI) part in \ref{dist2} follows. However, the results from \citeA{ChemlaCrnicFox:2015}, \citeA{Marty:2023} and \citeA{Degano:2023} show that 
 the inferences in \ref{Ig2} and \ref{dist2} are accessible to speakers even in the absence of the corresponding inferences in \ref{Ig1} and \ref{dist1}, suggesting that the former may arise independently of the latter. 
These findings challenge traditional implicature approaches to \textsc{II}, \textsc{DI} and \textsc{FC}. They are compatible, on the other hand, with more recent implicature approaches, as well as non-implicature approaches to these inferences, such as \citeA{aloni2022logic}. The latter are not committed to similarities between them and regular SIs and can further derive \textsc{POS} and \textsc{DI} without their \textsc{UNC} and \textsc{NU} counterparts. 
 %First, whether they are implicatures at all, given the large differences in robustness, and, second, if they are, how they are to be derived, given the challenges to the traditional approach. These findings are instead compatible with non-implicature approaches to free choice, distributive, and ignorance inferences, such as \cite{aloni2022logic}, which are not committed to similarities between them and regular implicatures, and can derive possibility and distributive inferences without their uncertainty and negated universal counterparts. 

\subsection{This Study}
We seek to address whether state-of-the-art LLMs predict the fine-grained inferences arising from plain and embedded disjunctions that we just described. Further, we address whether they exhibit the same pattern of results exhibited in the human data which distinguishes between different accounts of these inferences. LLMs have been tested on entailments \cite[among others]{wang2019superglue}, implicatures \cite{schuster-etal-2020-harnessing, li2021predicting, hu-etal-2022-predicting}, and presuppositions \cite{parrish-etal-2021-nope, jeretic-etal-2020-natural, sieker2023your, sravanthi2024pub}, but to our knowledge, the cases above have not been looked at yet. 
%To assess the abilities of LLMs, we address two questions. 
To this end, we compare LLMs' performance to human data from three experiments in \citeA{Marty:2021b}, three experiments in \citeA{Degano:2023} and two experiments in \citeA{Marty:2023} and assess the fit to human results. Further, we compare if LLM predictions align to the same theoretical predictions as human results, which were compared to the traditional implicature account (TIA), revised implicature account (RIA, \citeNP{bar2020free}), and non-implicature account (NIA, \citeNP{aloni2022logic}) in the three studies. In particular, in parallel to what \citeA{Marty:2021b} do with humans, we test whether the predictions of LLMs for FC, II, and DI inferences differ from those of \textsc{scalar implicatures}. In addition, following \citeA{Marty:2023} and \citeA{Degano:2023}, we test whether DI and POS inferences can be predicted also in the absence of the corresponding NU and UNC inferences. The theoretically motivated contrasts tested are summarized in Table~\ref{tab:predictions}.

Before moving on to the details of the study, we should emphasize that the theories above are designed to predict humans' linguistic behavior and, as such, they are of course not directly theories of LLM mechanics. Nonetheless, we think that using the same experimental paradigms as those used with humans in experimental pragmatics allows us to systematically check whether LLMs, as powerful statistical (fine-tuned) models trained on a lot of textual data, predict the inferences above in a ‘package’ with SIs, or distinguish between them in a way that aligns with humans, thus providing a novel perspective on the debate above.

%\sr{so we want to see if LLM are like humans or like the TA predicts???}
%\pt{would it make sense to spell out above already? otherwise, should we refer to forthcoming explanations?}, we look at model predictions in light of various theoretical predictions. 

\begin{table}[h!]
\begin{center}
\caption{Relevant predictions of the three accounts. `IMP' indicates whether the account predicts the inference in question to be an IMPlicature; `IND' indicates whether the account predicts \textsc{possibility} and \textsc{distributive} inferences to arise INDependently from the corresponding \textsc{uncertainty} and \textsc{negated universal} ones.}
\label{tab:predictions}
\vskip 0.12in
%\begin{tabular}{ c | c c c c c c}
% & \multicolumn{2}{c}{\textsc{Ignorance}} & \multicolumn{2}{c}{\textsc{Distributive}} & \multicolumn{2}{c}{\textsc{Free choice}}\\ 
%Theories & IMP & IND & IMP & IND & IMP & IND\\ 
%\midrule
%TIA & yes & no & yes & no & yes & no\\
%RIA & yes & yes & yes & yes & yes & yes\\
%NIA & no & yes & no & yes & no & yes\\
%C2015-M& - & - & 1 & 1\\
%\end{tabular}
\begin{tabular}{c|c |c c |c c}
 & \multicolumn{1}{c}{\textsc{Free choice}} & \multicolumn{2}{c}{\textsc{Ignorance}} & \multicolumn{2}{c}{\textsc{Distributive}}\\ 
Theories & IMP & IMP & IND & IMP & IND\\ 
\hline
TIA & yes & yes & no & yes & no\\
RIA & yes & yes & yes & yes & yes\\
NIA & no & no & yes & no & yes \\
%C2015-M& - & - & 1 & 1\\
\end{tabular}

\end{center}
\end{table}

\section{Experiments and Data}
%The goal of our experiments is to test whether and which pragmatic inferences are predicted by different large language models (LLMs), given sentences in context.
To assess LLM predictions, we closely replicated the human experiments' set-up of the \emph{mystery box} paradigm from \citeA{Degano:2023}, \citeA{Marty:2021b} and  \citeA{Marty:2023}. In this way, we were able to directly compare the humans' performance in these experiments with the LLMs' performance. %We compared the LLMs to human performance in these tasks and to theoretical predictions. Since LLM experiments closely replicated the human set-up, 
Therefore, we first describe the three human studies. 

\subsection{Human Studies}
\label{sec:human-expts}
In all experiments, participants were presented with pictures of three boxes whose contents were visible and one box whose content was not visible, the so-called \emph{mystery box}. The visible boxes contained one or two colored balls and the fourth mystery box had a question mark on it. Participants were instructed that the mystery box always has the same contents as one of the visible boxes. They were introduced with two child characters (Sam and Mia). The characters were familiarized with the rule about the mystery box and, therefore, they could make certain inferences about its contents. In each trial, the four boxes were presented and a sentence was uttered by one of the characters. The truth value of the sentences was manipulated by varying the contents of the visible boxes. Participants' task was a two-alternative forced choice task wherein they had to judge the sentences as ``good'' or ``bad'' given the context and the mystery box rule.
%\pt{The test sentences (\emph{trigger}) referred to the contents of the mystery box.}
%\pt{The rule in all experiments stated that the mystery box always contained the same contents as one of the three open boxes.

In \citeA{Marty:2021b} (abbreviated ``MRo''), all target trigger sentences were about the mystery box and tested the robustness of FC, DI and II inferences against the robustness of regular SIs. For our purposes, we selected materials from Experiments 4--6 of \citeA{Marty:2021b} which only contained positive polarity trigger sentences. For FC, the trigger sentence was of the form ``It is possible that the mystery box contains either a [A] ball or a [B] ball'', where [A] and [B] are placeholders for different color adjectives (e.g., yellow and blue); for DI, it was of the form ``It is certain that the mystery box contains either a [A] ball or a [B] ball''; for II, it was of the form ``The mystery box contains a [A] ball or a [B] ball''; finally, for SI, it was of the form ``It is possible that the mystery box contains a [A] ball''. In the critical trials, trigger sentences were presented in a \textsc{Target} context like the one in Table~\ref{tab:example} (first row), where every visible box contained an [A] ball and no [B] ball. \textsc{Target} contexts were designed so that the trigger sentence was false if the inference of interest is present, but true if it is absent. Thus, the more robust a given type of inference, the more participants should select the ``bad" response option in these trials and, consequently, the lower the acceptance rate should be.
There were 9 test trials per inference. %\pt{not sure how to best phrase this: A high trigger acceptance rate indicated that ...}

In \citeA{Degano:2023} (abbreviated ``D''), all target trigger sentences were about the mystery box and were of the form: ``The mystery box contains a [A] ball or a [B] ball''. There were two target contexts, \textsc{Target-1} and \textsc{Target-2}. \textsc{Target-1} contexts made \textsc{POS} inferences true and  \textsc{UNC} inferences false; \textsc{Target-2} contexts made both these inference types false (see Table~\ref{tab:example}, second and third row). As in MRo study, acceptance rates in these contexts were used as a proxy measure for the robustness of the inferences of interest: the lower the acceptance rate, the more robust the inference(s). \citeA{Degano:2023} found that \textsc{Target-1} contexts yield higher acceptance rates than \textsc{Target-2} contexts, suggesting that \textsc{POS} inferences are accessible independently of \textsc{UNC} inferences.
There were 36 test trials.

%\pt{All studies used the colors ... . Add condition description of Degano?}
In \citeA{Marty:2023} (abbreviated ``MRa''), target trigger sentences were either about the mystery box or about the visible boxes. The disjunction in them was embedded either under a \textsc{nominal} or a \textsc{modal} universal quantifier. For the \textsc{nominal} cases, the trigger sentence was of the form ``Every visible box contains a [A] ball or a [B] ball''. For the \textsc{modal} cases, the trigger sentence was of one of two forms: ``The mystery box must contain a A ball or a B ball'' (epistemic \textit{must}, Exp.~1) or ``[Name] must pick a [A] ball or a [B] ball'' (deontic \textit{must}, Exp.~2), where [Name] is a placeholder for a character's name. Target contexts in this study followed the same logic as in D study above and applied it to \textsc{DI} inferences: \textsc{Target-1} contexts made \textsc{NU} inferences true and  \textsc{DI} inferences false whereas \textsc{Target-2} contexts made both these inference types false. There were 24 test trials.

%The former type of sentence has the potential to give rise to ignorance inferences and the latter to distributivity inferences.   
%\pt{i refactor the example figure and added examples there (are they correct?) -- would something like this work?}
%\begin{itemize}
%    \item Target 1: The mystery box \{must contain, contains\} a blue ball or a yellow ball.
%    \item Target 2: The mystery box \{must contain, contains\} a blue ball or a yellow ball.
%\end{itemize}

%\sr{There were 6 repetitions of each target condition and 3 relations of 8 control conditions.} 
All three studies further contained \textsc{True} and \textsc{False} control contexts for each target trigger sentence. These contexts were similar in composition to the target ones but were designed so as to make the trigger sentences either true or false  independent of the target inference. Control contexts served to provide clear baselines for acceptance and rejection of the target sentences under investigation. Additionally, all three studies included control trials involving non-target trigger sentences associated with true and false contexts. Responses to these trials were used to assess participants' general performance in the task independent of the critical items. In total, there were 72, 108 and 72 control trials, respectively.

%\subsection{Hypotheses}
%\pt{spell out predictions for the single experiments? not sure if it needs to be a separate subsection, though. Also I saw that @SR is already doing this by-human-study, great.}

\subsection{Experiments with LLMs}
\begin{figure*}[ht]
\begin{center}
\includegraphics[scale = 0.35]{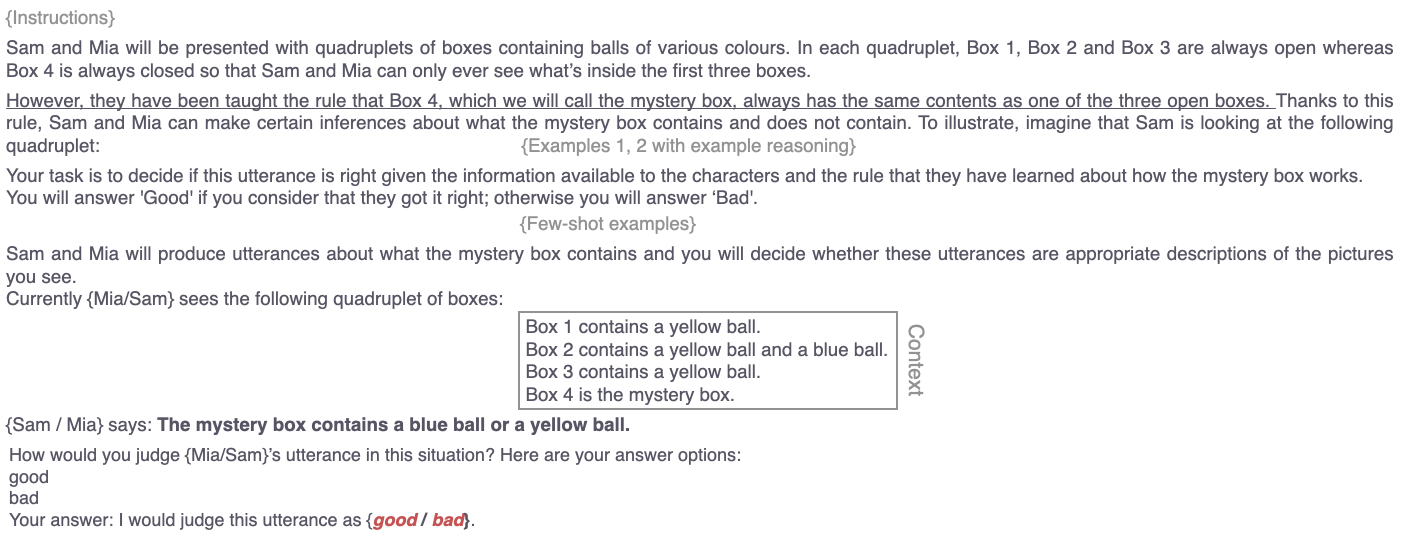}
\end{center}
\caption{Example prompt for the LLM experiments. Some parts of the prompt are omitted for brevity (in gray). Boldface trigger sentence is an example from \citeA{Degano:2023}. Underlined sentence is the mystery box rule. Expressions in curly braces in gray vary by study. The character name is sampled at random by-trial. The likelihood for the last word (one of ``good'' / ``bad'', italicized) is retrieved for scoring the trigger, given the context.} 
\label{fig:item-example}
\end{figure*}

%\subsection{Materials}
%In order to compare the LLM predictions to human data, we draw on the experimental materials by \citeA{Degano:2023, Marty:2021b, Marty:2023}. 
Materials for the LLM studies were constructed by converting all selected vignettes from the human studies described in the previous section to text-based prompts for the LLMs. That is, the visual stimuli of the boxes were converted to textual descriptions. For instance, the \textsc{Target-1} context from Table~\ref{tab:example} was represented as shown in Figure~\ref{fig:item-example}. General instructions, the cover story, the mystery box rule and examples were prepended to the critical context.
%All human experiments were forced choice experiments where participants had to indicate whether they accept a trigger sentence in a particular context by choosing between the answer options ``good'' and ``bad''. 
An instruction presenting the answer options ``good'' and ``bad'' in randomized order was added to each prompt (see Fig.~\ref{fig:item-example}).\footnote{Full materials can be found under: \url{http://tinyurl.com/4asyhzvv}}

\begin{table}[h!]
\begin{center}
\caption{Example \textsc{target} context in \citeA{Marty:2021b} and example \textsc{target-1} and \textsc{target-2} contexts in \citeA{Degano:2023} and \citeA{Marty:2023}. In these examples, [A] is yellow and [B] is blue.}
\label{tab:example}
\vskip 0.12in
\begin{tabular}{lcccc}
Context & \multicolumn{4}{l}{Example}\\
\toprule
\textsc{Target} 
& \makecell{\includegraphics[scale=0.1]{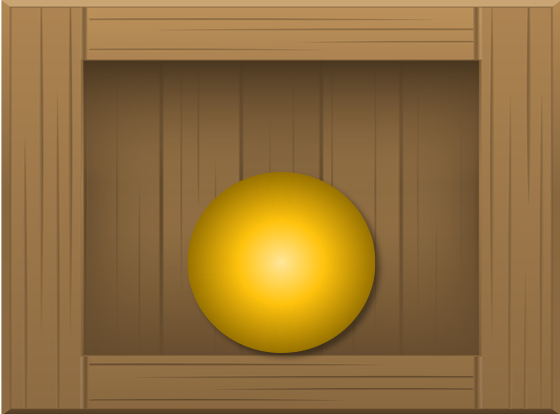}} & 
\makecell{\includegraphics[scale=0.1]{yw.png}} &
\makecell{\includegraphics[scale=0.1]{yw.png}} &
\makecell{\includegraphics[scale=0.1]{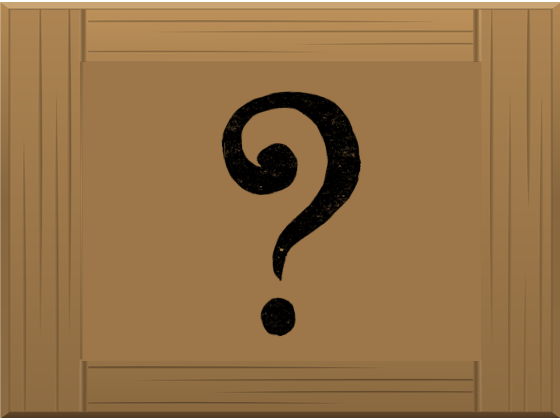}}  \\
& A & A & A & ? \\
\textsc{Target-1} 
& \makecell{\includegraphics[scale=0.1]{yw.png}} & 
\makecell{\includegraphics[scale=0.1]{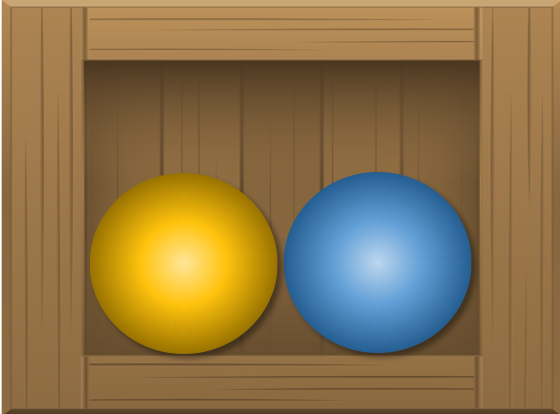}} &
\makecell{\includegraphics[scale=0.1]{yw.png}} &
\makecell{\includegraphics[scale=0.1]{mystery.png}} \\
& A & AB & A & ?  \\ 
\textsc{Target-2} 
& \makecell{\includegraphics[scale=0.1]{yw.png}} & 
\makecell{\includegraphics[scale=0.1]{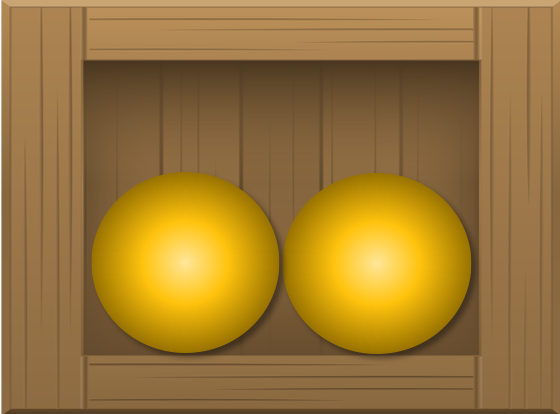}} &
\makecell{\includegraphics[scale=0.1]{yw.png}} &
\makecell{\includegraphics[scale=0.1]{mystery.png}}  \\
& A & AA & A & ?
%\midrule
%\multicolumn{5}{l}{}
%\textit{`The mystery box contains a yellow}}\\
%\multicolumn{5}{l}{\textit{ball or a blue ball.'} \hfill $(A \lor B)$}
\end{tabular}
\end{center}
\end{table}

\subsection{Methods}
We tested a range of LLMs where retrieval of log probabilities of strings is possible, selecting different state-of-the-art models so as to cover models with different architectures, fine-tuning, training data composition and scale. Specifically, we used the following LLMs: \texttt{text-davinci-003} version of GPT-3.5 \cite{NEURIPS2020_1457c0d6}, Llama-2 (7B, 13B, 70B parameters, base and chat versions, abbreviated ``L-Xb'') \cite{touvron2023llama}, Mistral-7B (v0.1) and Mistral-7B-Instruct (v0.2) \cite{jiang2023mistral}, Mixtral-8x7B (v0.1) and Mixtral-8x7B-Instruct (v0.1) \cite{mistralai2023mixtral}, Pythia (2.8B, 6.9B and 12B parameters, \citeA{biderman2023pythia}), Phi-2 \cite{li2023textbooks} and Falcon-7B \cite{falcon40b}.\footnote{Except for GPT-3.5, all models are open-access. At the time of submission, the specific model endpoint was discontinued by the provider OpenAI.} 

We used the same prompting and scoring strategies for retrieving predictions from all models for all experiments. Specifically, we follow the design of human studies previously described. % Section~\ref{sec:human-expts}. 
Human subjects completed training trials where they learned the experimental task by seeing control conditions and receiving feedback about correctness of their responses. We use all control trials with correct answers that were used in the human training phase as a few-shot prompt for each main trial for the LLMs. 
Therefore, each experimental item $i$ consisted of a prompt $C_i$ (including the instructions and the cover story description, the few-shot prompt, the critical item context and the task) and the two answer options $o_{j}=t_{j1}\dots t_{jn}$, $j \in \{\text{``good'', ``bad''}\}, n\geq 1$, each consisting of $n$ tokens (depending on the model's tokenizer).

We computed an LLM prediction $S_i$ for item $i$ via retrieving the log probability $\log P_{LLM}(o_{j} \mid C_i)$ of each $o_{j}$ following the item prompt, under each LLM, respectively. We used the average token log probability if an answer option consisted of multiple tokens (cf.~\citeA{HoltzmanWest2021:Surface-Form-Co}): 
\begin{equation}
    \log P_{LLM}(o_{j} \mid C_i) = \frac{1}{n}\sum_{l=1}^n log P_{LLM}(o_{jl} \mid C_i, o_{j<l})
\end{equation}
Given the scores for the two answer options, we identified the LLM prediction for a given item $i$ as the answer option with the maximal log probability $S_i = \arg\max_j \log P(o_{j} \mid C_i)$.

For control trials where a correct answer existed, we computed \emph{accuracy} of LLMs (i.e., the proportion of items where the chosen response option was correct). For target conditions, we computed the \emph{acceptance rate} predicted by LLMs as the proportion of trigger sentences for which the option ``good'' was chosen (i.e., $\log P_{LLM}(\text{good} \mid C_i) > \log P_{LLM}(\text{bad} \mid C_i)$). The predicted acceptance rate was used to assess the closeness of LLM predictions and human results. Specifically, we computed the proportion of variance in human data explained by the model predictions by calculating $R^2$ for the simple linear model regressing human acceptance rates against model predictions, for each condition of each experiment. To identify overall model fit to human results across experiments, we calculated the adjusted $R^2$ \cite{miles2005r}.\footnote{The linear model was \texttt{human\_accept.rate $\sim$ model\_accept.rate + condition}, with seven distinct conditions from the three replicated studies.}

%\pt{potentially address correlation of control accuracy with critical trial results}
\begin{table*}[!ht]
    \begin{center}
    \caption{Average accuracy on control items by source. Boldface indicates highest accuracy among LLMs.} 
    \label{tab:control-accuracy}
    \vskip 0.12in
    \begin{tabular}{|c|cccccc|}
    \hline
               Study & Human & GPT-3.5 & Llama-2-70b & Mixtral-Instruct & Mixtral & Mistral-Instruct\\ \hline
        MRo & 0.89 & \textbf{0.94} & 0.69 & 0.74 & 0.65 & 0.54\\ \hline
        D & 0.95 & 0.72 & \textbf{0.9} & 0.75 & 0.53 & 0.64\\ \hline
        MRa & 0.94 & 0.76 & \textbf{0.86} & 0.71 & 0.82 & 0.75\\ \hline
    \end{tabular}
    \end{center}
\end{table*} %

\begin{figure*}[ht]
\begin{center}
\includegraphics[scale = 0.58]{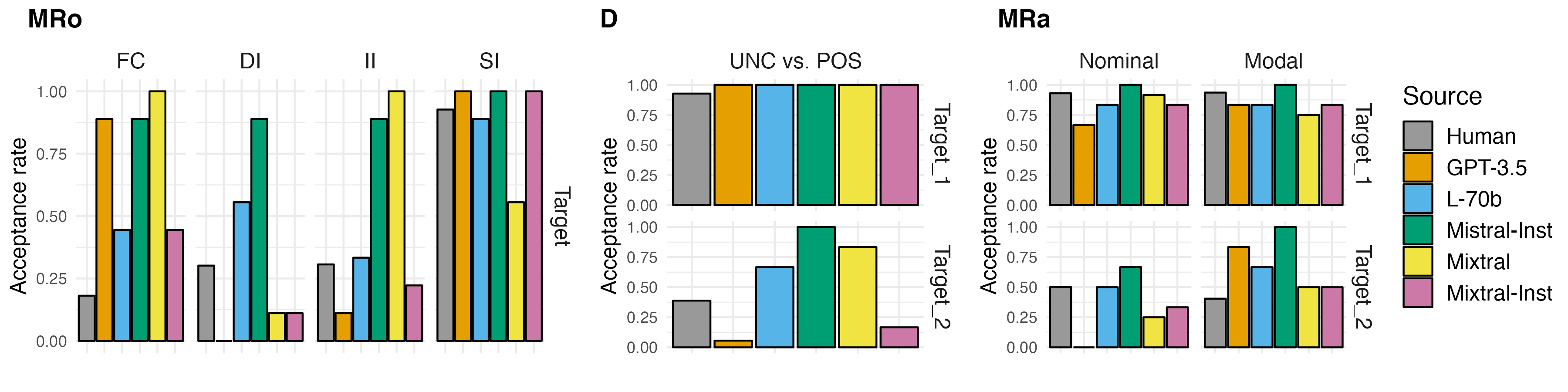}
\end{center}
\vspace{-5ex}
\caption{Mean acceptance rate in the target conditions of each study by test case and source (LLMs or humans).} 
\label{fig:results-acc-rate}
\end{figure*} %

\begin{figure*}[!ht]
\begin{center}
\includegraphics[scale = 0.58]{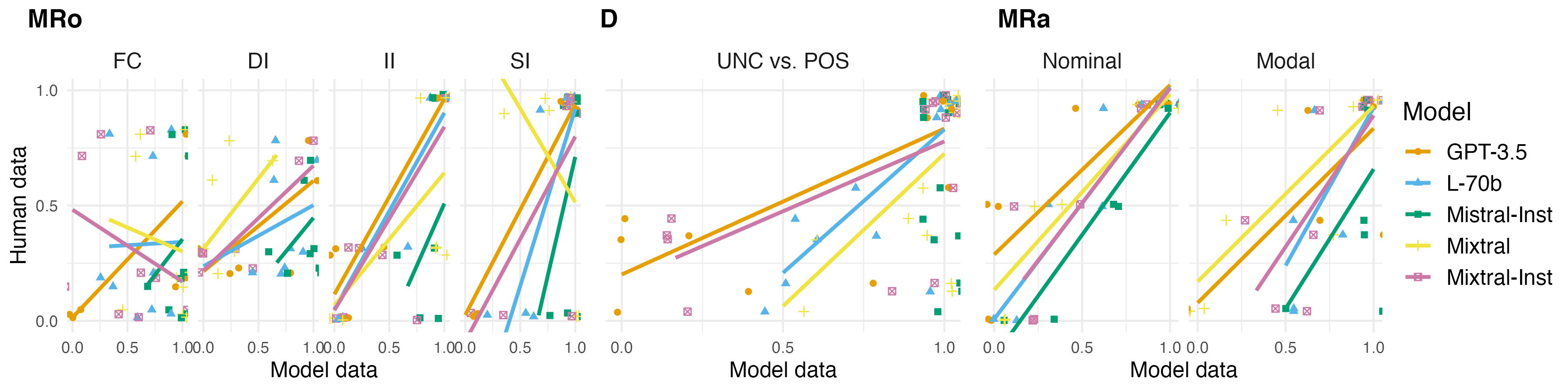}
\end{center}
\vspace{-4ex}
\caption{Human acceptance rate averaged over all items in each condition, by trigger type, plotted against model predictions (points). Lines indicate best linear model fit regressing human data against model predictions.} 
\label{fig:results-lm}
\end{figure*} %

\begin{table*}[!ht]
    \begin{center}
    \caption{Goodness of fit of model predictions to human results in each study and for each test case, reported as the $R^2$ value for \texttt{human\_data $\sim$ model\_data}. Overall model fit reports adjusted $R^2$ with 95\% confidence intervals (estimated using percentiles
from 10,000 bootstrap samples). Higher values are better. Boldface indicates the best model(s) in each condition.}
    \label{tab:human-fit-r2}
    \vskip 0.12in
    \begin{tabular}{|c|c|ccccc|}
    \hline
        Study & Case & GPT-3.5 & Llama-2-70b  & Mistral-Inst. & Mixtral & Mixtral-Inst. \\ \hline
        \multirow{4}{*}{MRo} & FC & 0.53 & 0 & 0.1 & 0.1 & \textbf{0.75} \\ %\hline     
            & DI & 0.65 & \textbf{0.97} & \textbf{0.97} & \textbf{0.97} & \textbf{0.97} \\ %\hline
            & II & 0.95 & \textbf{0.99} & 0.9 & 0.44 & 0.9 \\ %\hline
            & SI & \textbf{0.99} & 0.95  & \textbf{0.99} & 0.22 & \textbf{0.99} \\ \hline
        D & UNC vs.\ POS & 0.71 & \textbf{0.91}  & 0 & 0.9 & 0.49 \\ \hline
        \multirow{2}{*}{MRa} & Nominal & 0.73 & \textbf{0.99} & \textbf{0.99} &
        0.92 & 0.88 \\ %\hline
          & Modal & 0.73 & 0.9  & 0.66 & \textbf{0.93} & 0.7 \\ \hline
        \multicolumn{2}{|r}{\textit{Overall model fit:}}  & 0.68 [0.60, 0.92] & 0.66 [0.50, 0.94] & 0.23 [0.18, 0.78]  & 0.18 [-0.04, 0.86] & 0.42 [0.23,0.88] \\ \hline
        %\multicolumn{2}{|r}{\textit{Bootstrapped 95\% CIs:}}  &  &  & &  &  \\ \hline
    \end{tabular}
    \end{center}
\end{table*}

\subsection{Results}
We first compute the accuracy of all models on the control trials of each study. We then take the top five models with the highest accuracy and further investigate the performance of only those models on critical trials. The control accuracy scores of these selected models are reported in Table~\ref{tab:control-accuracy}. Full results for all models can be found in the repository.
This selection strategy of best models deviated from the design of human studies where the exclusion criterion for participants was an accuracy of 0.8 in \citeA{Degano:2023} and \citeA{Marty:2023} and 0.7 in \citeA{Marty:2021b}.
Table~\ref{tab:control-accuracy} indicates that the performance of LLMs was above chance (0.5 for a single trial), but it was not consistent across the studies. Furthermore, with one exception, LLMs performed worse than humans.

Second, we investigate the robustness of inferences predicted by the LLMs based on triggers with disjunctions in different configurations, and whether the predicted acceptance rates were similar to human inference rates. To this end, we compared the acceptance rates of the target triggers in the \textsc{FC}, \textsc{II}, \textsc{DI} conditions to the acceptance rate of the \textsc{SI} condition (Fig.~\ref{fig:results-acc-rate}, left facet, first three facets~vs.~last facet). Visual inspection suggests that some models robustly predicted \textsc{FC, DI, II} inferences but not \textsc{SIs} (L-70b, Mixtral-Inst.), while other models were inconsistent with respect to which inferences they predicted. % \pt{If we wanted to go for the ``pragmatic inference package framing, should this be added here or in the discussion?''}
Next, we assess the LLMs' fit to human responses via $R^2$. To this end, we considered LLM predictions on both target trigger conditions and control conditions (acceptance rates for controls are not shown).
Comparing LLM predictions to human inferences, we found that for different inferences, different models best captured human data in terms of explained variance (see Table~\ref{tab:human-fit-r2}, MRo results). Figure~\ref{fig:results-lm} (left) indicates that LLMs were less consistent across triggers with respect to human data in the \textsc{FC} and \textsc{DI} conditions, than in the other two conditions.
%Overall, these results indicate that predictions of some models would challenge the TIA and RIA accounts of \textsc{FC, DI, II} inferences and are rather consistent with NIA given the varying acceptance rates \jr{maybe let's leave this to the discussion with the caveat that those theories were not meant to be theories of LLMs but if we apply them to those ...} (cf.~Table~\ref{tab:predictions}), while other models' predictions remain difficult to interpret.

Third, we turn to the question of whether LLMs predict that \textsc{POS} and \textsc{DI} occur together with their corresponding \textsc{UNC} and \textsc{NU} inferences. Figure~\ref{fig:results-acc-rate} (middle) shows the mean acceptance rates for the experiment wherein the triggers contained plain disjunctions, investigating \textsc{II}. Visually, the crucial comparison between the \textsc{Target-1} and \textsc{Target-2} conditions was borne out in prediction of GPT-3.5, L-70b and both Mixtral models, albeit less for the base version. Comparing the overall fit of model predictions to human data across triggers, we found that L-70b best captured human responses (Table~\ref{tab:human-fit-r2}, D). Figure~\ref{fig:results-lm} indicates that, both L-70b and Mixtral performed close to human data, while Mistral-Instruct did poorly by always accepting both Target conditions. %Predictions of all but Mistral models are consistent with the RIA and NIA accounts (cf.~Table~\ref{tab:predictions}). 

Figure~\ref{fig:results-acc-rate} (right) shows the mean acceptance rates for the experiment wherein the triggers contained a disjunction embedded under a nominal quantifier or a modal, investigating \textsc{distributive} inferences. 
While %\jr{I mean not only visually we showed in Marty et al that the pattern is the same with no interaction across modals and nominals. or did you mean "visually" for LLMs? in that case we can move it later ...e.g. , where, at least visually, LLM predictions varied across contexts } 
human inferences remained largely unaffected across modal and nominal contexts (leftmost bars, left~vs.~right facets; cf.~\citeA{Marty:2023}), at least visually, LLM predictions varied more across contexts (left~vs.~right facets). Focusing on the critical contrast between the \textsc{Target-1} and \textsc{Target-2} conditions, we found that it was borne out in predictions of all models in the nominal context, but not for GPT-3.5 and Mistral-Instruct in the modal context. For other models, the contrast was also smaller in the modal context. This discrepancy was reflected in the fit to human data (see Table~\ref{tab:human-fit-r2}, MRa rows, lower for modal than for nominal cases for some models). The modal context also appears to be more noisy than the nominal context in Figure~\ref{fig:results-lm}.
%Overall, models which predicted the difference between the Target-1 and Target-2 conditions across context provide results consistent with the RIA and NIA accounts (cf.~Table~\ref{tab:predictions}). 

To summarize the comparison of LLMs to human predictions across studies and conditions, we found that, overall, GPT-3.5 explained the most variance in human data, closely followed by L-70b (Table~\ref{tab:human-fit-r2}, last row). However, the fit to human predictions is not perfect and varies depending on the test conditions, even for the best-performing models. While L-70b did poorly on FC, it was otherwise better than GPT-3.5 for many other conditions.

\section{Discussion}
In this paper, we set out to investigate whether LLM evaluations can shed light onto theoretical debates about the status of different linguistic inferences.
In a concrete case study, we compared the predictions of state-of-the-art LLMs with the main inferences of plain and embedded disjunctions, \textsc{FC}, \textsc{DI} and \textsc{II} in comparison with \textsc{SIs}, with the human results of \citeA{Marty:2021b, Marty:2023} and \citeA{Degano:2023}. In our results, we find a clear reflection of the large difference found in humans between regular \textsc{SIs} and \textsc{FC}, \textsc{II}, and \textsc{DI} in the best performing models we tested. 
However, future experiments could extend the results, e.g., regarding the ability of LLMs to consistently perform on sentences with modals, as well as with negation (cf.~\citeA{Marty:2021b}), as it has been found that LLMs might have issues with negation \cite{truong2023language}. Furthermore, LLMs might be sensitive to superficial aspects of the prompts \cite{liu2023lost}, which we would not expect to affect human performance; therefore, testing the robustness of predictions on pragmatic inferences under different prompting strategies should be further analyzed.

%Overall, we found that LLMs produced different predictions for various types of pragmatic inferences. 
%Overall, the LLM results show that the best performing models mostly align with humans in the fine-grained distinctions between the inferences above and regular implicatures, as well as on different aspects of those inferences. 
Overall, the results for many conditions, like those for humans, are not in line with traditional implicature approaches to those inferences, i.e., showing that \textsc{POS} and \textsc{DI} can be present in the absence of their \textsc{UNC} and \textsc{NU} counterparts. 
However, different LLMs still exhibit different noticeable inconsistencies across conditions in non human-like ways (e.g., \textsc{FC}~vs.~\textsc{SI} rates of GPT-3.5). 
Therefore, results of the present study should be taken with a grain of salt in context of the theoretical debate about the status of these inferences.
Nonetheless, we use our results as an opportunity to suggest some criteria that might be required in order to draw robust insights from LLMs for linguistic theories.
We hypothesize that results from LLMs might be more informative as the performance becomes more human-like. 
For this, first, it is critical that LLMs perform consistently on a suite of tasks testing closely related phenomena (e.g., akin to the presented set of inferences). %This would decrease the likelihood of over-reliance on results that might be due to potential spurious correlations rather than target capabilities. 
Second, it is important to assess the whole distribution of LLM responses (including potential error analysis), and not just focus on one target condition. Performance on control conditions should also be taken into account. Further, with increased interpretability, LLM results might become more trust-worthy. Finally, as suggested by \citeA{hu2024auxiliary}, performance of LLMs of different capacity (e.g., number of parameters, training data size) should be compared to human acquisition trajectories. If performance scales with LLM capacity according to task and phenomenon complexity in a human-like way, this would further justify taking LLM performance as evidence bearing on theoretical debates. Our results preliminarily suggest that model size is a predictor of fit to human data on these inferences (cf.~Mistral results~vs.~other models). 
In sum, our study's contributions are twofold: first, we show that a complex experimental paradigm used in human linguistic experiments, the mystery box paradigm, can be used with LLMs; second, we show that systematic comparison of human results and LLM performance can contribute fruitful insights on the relation between LLM results and linguistic theory, a debate which will likely become more important in the next years.

%\pt{effects of prompting, scaling, fine-tuning, the usual. Interestingly, the chat llama models were much worse than the base ones. Our zero-shot explorations can be mentioned. Outlook to more fine-grained comparisons if not verbalized above (by-experiment, negation for \citeA{Marty:2021b} etc)}

\section{Acknowledgments}
We would like to thank Todd Snider and the anonymous reviewers for insightful feedback. We gratefully acknowledge support by the state of Baden-W\"urttemberg, Germany, through the computing resources provided by bwHPC and the German Research Foundation (DFG) through grant INST 35/1597-1 FUGG. MF is a member of the Machine Learning Cluster of Excellence, EXC number 2064/1 – Project number 39072764. SR was supported by the NWO OC project Nothing is Logical (grant no 406.21.CTW.023).

\bibliographystyle{apacite}

\setlength{\bibleftmargin}{.125in}
\setlength{\bibindent}{-\bibleftmargin}

\bibliography{CogSci_Template}

\end{document}